%% file: main.tex
\documentclass{bmvc2k}

\title{GTA: Global Temporal Attention for Video Action Understanding}

\addauthor{Bo He$^*$}{bohe@umd.edu}{1}
\addauthor{Xitong Yang$^*$}{xyang35@cs.umd.edu}{1}
\addauthor{Zuxuan Wu}{zxwu@fudan.edu.cn}{2}
\addauthor{Hao Chen}{chenh@umd.umd}{1}
\addauthor{Ser-Nam Lim}{sernamlim@fb.com}{3}
\addauthor{Abhinav Shrivastava}{abhinav@cs.umd.edu}{1}

\addinstitution{
    University of Maryland, \\
    College Park, MD, USA
}
\addinstitution{
    Fudan University, \\
    Shanghai, China
}
\addinstitution{
    Facebook AI, \\
    Sunnyvale, CA, USA
}

\runninghead{HE ET AL.}{Global Temporal Attention for Video Action Understanding}

\usepackage{times}
\usepackage{epsfig}
\usepackage{graphicx}
\usepackage{amsmath}
\usepackage{amssymb}

\usepackage{footmisc}
\usepackage{booktabs}
\usepackage{multirow}
\usepackage{adjustbox}
\usepackage{mathtools}

\usepackage{floatrow}
\usepackage{hyphenat}
\newfloatcommand{capbtabbox}{table}[][\FBwidth]
\DeclareFloatSeparators{myfill}{\hskip.013\textwidth plus1fill}

\usepackage[section]{placeins}

\def\eg{\emph{e.g}\bmvaOneDot}

\def\ie{\emph{i.e}\bmvaOneDot}

\DeclarePairedDelimiter\floor{\lfloor}{\rfloor}

\newcommand{\system}{GTA\xspace}

\begin{document}

\maketitle

\vspace{-0.2in}
\begin{abstract}
Self-attention learns pairwise interactions to model long-range dependencies, yielding great improvements for video action recognition. In this paper, we seek a deeper understanding of self-attention for temporal modeling in videos. We first demonstrate that the entangled modeling of spatio-temporal information by flattening all pixels is sub-optimal, failing to capture temporal relationships among frames explicitly. To this end, we introduce Global Temporal Attention (GTA), which performs global temporal attention on top of spatial attention in a decoupled manner. We apply \system on both pixels and semantically similar regions to capture temporal relationships at different levels of spatial granularity. Unlike conventional self-attention that computes an instance-specific attention matrix, \system directly learns a global attention matrix that is intended to encode temporal structures that generalize across different samples. We further augment \system with a cross-channel multi-head fashion to exploit channel interactions for better temporal modeling. Extensive experiments on 2D and 3D networks demonstrate that our approach consistently enhances temporal modeling and provides state-of-the-art performance on three video action recognition datasets.
\end{abstract}

\input{introduction}

\input{relatedwork}

\input{approach}
\input{experiment}

\vspace{-0.1in}
\section{Conclusion}
\vspace{-0.05in}
In this paper, we present Global Temporal Attention (\system), which is designed for improved temporal modeling in video tasks. \system is built upon a decoupled self-attention framework, where temporal attention is disentangled from the spatial attention to prevent being dominated by the spatial one. We apply \system to model the temporal relationships at both pixel-level and region-level. Moreover, \system directly learns a global, instance-independent attention matrix that generalizes well across different samples. A cross-channel multi-head mechanism is also designed to further improve the temporal modeling in \system.
Experimental results demonstrate that our proposed \system effectively enhances temporal modeling and achieves state-of-the-art results on three challenging video action benchmarks.

\paragraph{Acknowledgements.} This work was supported by the Air Force (STTR awards \\FA865019P6014,  FA864920C0010), DARPA SemaFor program (HR001120C0124), and an independent gift from Facebook AI.

\bibliography{egbib}

\clearpage
\input{appendix}

\end{document}

%% file: introduction.tex
\section{Introduction}
\begin{figure}[!htb]
    \centering
    \includegraphics[width=0.7\linewidth]{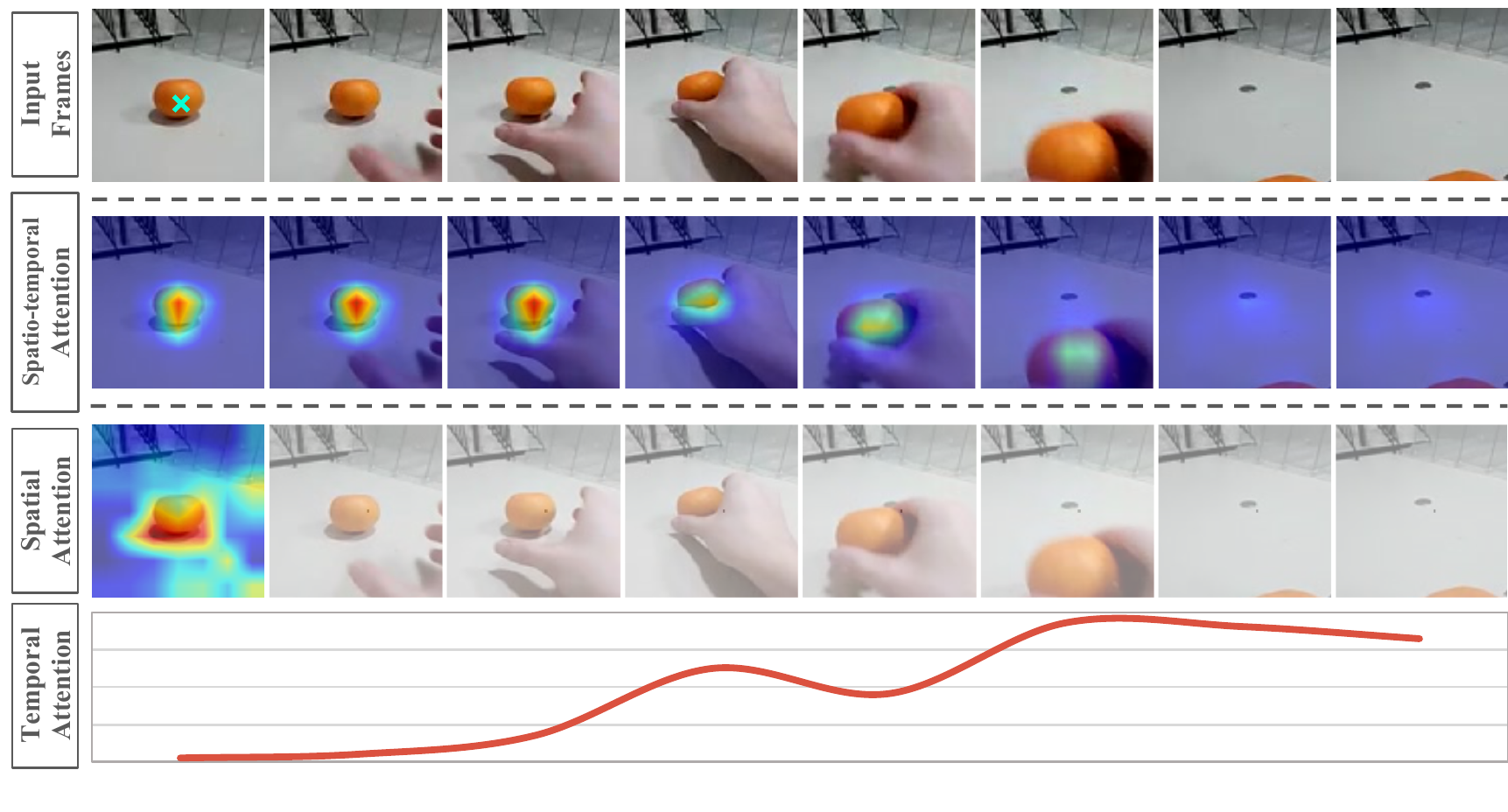}
    \caption{\textbf{Top}: input frames (action: \textit{removing something to reveal something}). The green cross-mark indicates the query position. \textbf{Center}: spatio-temporal attention generated by NL blocks. The attention is biased towards the appearance similarity, which fades overtime ignoring temporal clues; thus, the model generates incorrect prediction: \textit{putting something in front of something}. \textbf{Bottom}: the decoupled NL blocks generate spatial attention maps within the query frame and temporal attention weights across different time steps. The temporal attention has larger values at the key frames that are critical for recognizing the action (\ie, revealing something), and the model gives the correct prediction. \system is built upon the decoupled framework and advances the temporal attention to a more effective design.\vspace{-0.1in}}
    \label{fig:teaser}
\end{figure}

Attention mechanisms have demonstrated impressive achievements in a wide range of tasks such as language modeling~\cite{bahdanau2014neural,vaswani2017attention}, speech recognition~\cite{chorowski2015attention} and image classification~\cite{hu2018squeeze,Hu2018GatherExciteEF}. 
One of the most effective attention methods is self-attention, which learns self-alignment via dot product operations, computing pairwise similarities between a pixel (\ie, query) and other pixels (\ie, key) to modulate the transformed inputs (\ie, value).
For action recognition~\cite{Wang2017NonlocalNN}, this requires: (i) flattening all pixels in a video, regardless of their spatial and temporal locations, into a huge vector; (ii) sharing the same set of parameters for all pixels to derive the query/key/value; and (iii) generating a joint attention map for both spatial and temporal context.

In this paper, we seek a better understanding of self-attention for temporal modeling in videos. In particular, we wish to answer the following questions: (i) Is treating all pixels in space and time as a flattened vector to perform dot-product sufficient for temporal modeling? (ii) Is dot product based self-attention really necessary for capturing temporal relationships across different frames?

In contrast to the conventional use of self-attention for video recognition, we posit that temporal attention should be \textit{disentangled} from spatial attention, since they focus on different aspects. As shown in Figure~\ref{fig:teaser}, the spatial attention tends to capture appearance similarity (\ie, the orange), while the temporal attention is more focused on frames that are important for recognizing the action (\ie, revealing something). When these two types of attention are modeled together (Figure~\ref{fig:teaser} Center), the attention is biased towards the appearance similarity, dominating any temporal context.

In addition, we argue that dot product based self-attention is \emph{not even suitable} for temporal modeling. Standard self-attention produces instance-specific attention weights, conditioned on pairwise interactions. In the spatial domain, it can attend to salient regions for improved performance. 
When used for temporal modeling, it ignores the ordering of frames as self-attention is known to be permutation invariant~\cite{DBLP:conf/iclr/CordonnierLJ20}. For instance, if we shuffle two pixels temporally, their relationship will be the same, producing the same output. This is not sufficient for differentiating actions like ``reveal something'' and ``cover something''. We hypothesize that temporal modeling requires learning a \emph{global} temporal structure that generalizes across different samples rather than relying on pairwise interactions across time steps.

In light of this, we introduce Global Temporal Attention (\system), for video action recognition. In particular, we first decouple the traditional spatio-temporal self-attention into two successive steps---a standard self-attention in the spatial domain within each frame followed by the proposed \system module to capture temporal relationships across different frames.
Moreover, we not only apply \system to each pixel location along the temporal dimension but also ``superpixels''---pixels in a region share similar semantic meanings. This enables our model to capture temporal relationships at different levels of spatial granularity. Unlike computing pairwise frame interactions with dot product, \system directly learns a global attention matrix that is randomly initialized to be instance-independent. The intuition of the global attention matrix is to not rely on pairwise frame relations without specific ordering information or individual sample information, but to learn a global task-specific weight matrix considering temporal structures that generalize across different samples. To exploit information across different channels, we split feature maps into multiple groups along the channel-dimension, and for each group we apply \system in a multi-head fashion such that each head focuses on different aspects of the inputs. Then, outputs from different channel groups are further aggregated to produce a unified representation.

We conduct extensive experiments on Something-Something~\cite{goyal2017something} and Kinetics-400~\cite{kay2017kinetics}. Our proposed \system outperforms the traditional spatio-temporal self-attention by clear margins, and achieves state-of-the-art results on these three datasets. We also provide a side-by-side comparison with recent NL variants~\cite{yue2018compact,cao2019gcnet,chen2019graph} to show the superior performance of \system in temporal modeling.
We summarize our main contributions as follows.
First, we provide an in-depth analysis of the sub-optimal design of the spatio-temporal self-attention and propose to decouple attention across the two dimensions.
Second, we introduce \system, which improves the conventional temporal attention by introducing: (i) temporal modeling at both pixel and region levels; (ii) a global attention matrix for all samples; (iii) a cross-channel multi-head design for incorporating channel interactions.

%% file: relatedwork.tex
\vspace{-0.05in}
\section{Related Work}

\paragraph{Temporal Modeling in Action Recognition.}
A large family of research in action recognition focuses on the effective modeling of temporal information in videos.
Early work simply aggregates the frame/clip-level features across time via average pooling~\cite{karpathy2014large,wang2016temporal} or feature encoding like ActionVLAD~\cite{girdhar2017actionvlad}, without considering the temporal relationships of video frames.
Later on, two-stream networks~\cite{simonyan2014twostream}, 3D convolution networks (CNNs)~\cite{ji20123d,tran2015learning} and recurrent neural networks (RNNs)~\cite{Donahue2015LongtermRC,yue2015beyond} are used to model the spatial and temporal context in videos.
Recently, various temporal modules are proposed to capture temporal relations, such as TRN~\cite{zhou2018temporal} based on relation networks, Timeception~\cite{hussein2019timeception} based on multi-scale temporal convolutions, and SlowFast~\cite{feichtenhofer2019slowfast} based on slow and fast branches capturing spatial and motion information, respectively.
TSM~\cite{lin2019tsm} adopts a channel shifting operation along the time dimension to enable temporal modeling on 2D CNN networks. %
STM~\cite{jiang2019stm}, TEA~\cite{li2020tea} and MSNet~\cite{kwon2020motionsqueeze} encode the motion information into the network by extracting motion features between adjacent frames.

\vspace{-0.1in}
\paragraph{Non-Local and Self-Attention.}
Modeling long-range relations in feature representations has a long history~\cite{strat1991context,buades2005non,efros1999texture,gupta2015exploring,heitz2008learning,mottaghi2014role} and has proven to be effective in various tasks, such as machine translation~\cite{vaswani2017attention}, generative modeling~\cite{zhang2018selfattention},
image recognition\cite{Hu2018GatherExciteEF,Wang2017NonlocalNN,cao2019gcnet}, object detection\cite{hu2018relation,Wang2017NonlocalNN,cao2019gcnet}, semantic segmentation\cite{zhang2018context,Wang2017NonlocalNN,cao2019gcnet} and visual question answering\cite{Santoro2017ASN}.
In computer vision, Non-local Network (NL)~\cite{Wang2017NonlocalNN} is proposed to model the pixel-level pairwise similarities to encode long-range dependencies. 
SENet~\cite{hu2018squeeze} uses a Squeeze-and-Excitation block to model inter-dependencies along the channel dimension.
GCNet~\cite{cao2019gcnet}, CGNL~\cite{yue2018compact} and DANet~\cite{Fu_2019} further improve the vanilla NL by integrating pixel-wise and channel-wise attention. CCNet~\cite{huang2019ccnet} improves the efficiency of NL by computing the contextual information of the pixels on its crisscross path instead of the global region. GloRe~\cite{chen2019graph} proposes the relation reasoning via graph convolution on a region-based graph in the interaction space to capture the global information.

In this work, we present a novel way to model temporal relationships and bring new perspectives for a better understanding of the attention mechanism utilized in video action recognition.
Our approach learns global temporal attention that generalizes well across different samples as opposed to using pairwise interactions with dot product in self-attention.

%% file: approach.tex
\section{Approach}

\subsection{Background}
\label{sec:background}
Extending the self-attention module~\cite{vaswani2017attention} for language tasks, the non-local block (NL)~\cite{Wang2017NonlocalNN} takes as input flattened pixels in spacetime to model pairwise interactions, as shown in Figure~\ref{fig:model}(a). More formally, given an input feature map $X \in \mathbb{R}^{N \times C}$, three linear projections are applied to obtain key ($K$), query ($Q$), and value ($V$) representations, where $C$ is the channel dimension of the feature map. We use $N=THW$ to denote the total number of positions in both space and time dimensions, where $T$, $H$ and $W$ are the number of time steps, height and width of the feature map, respectively.
The three projections can be written as $Q = XW_Q, \  K = XW_K, \ V = XW_V$, parameterized by three weight matrices $W_Q, W_K, W_V \in \mathbb{R}^{C\times C}$ respectively.
The output of the self-attention operation is computed as a weighted sum
of the value representations. Here, the weight is defined by the attention weight matrix $M \in \mathbb{R}^{N\times N}$, where each element denotes a scaled dot product between the query pixel and the corresponding key pixel, followed by a softmax normalization:

\begin{equation}
   A = MV, \quad M =  \texttt{softmax}\left(\frac{QK^T}{\sqrt{C}}\right). \label{eq:dotproduct}
\end{equation}
The attention output is incorporated into the backbone network via a final linear projection $W^O \in \mathbb{R}^{C\times C}$ and a residual connection~\cite{he2016deep}:
\begin{equation}
\label{eq:output}
    Y = X + AW^O,
\end{equation}
An optional normalization layer (\eg, BatchNorm~\cite{ioffe2015batch} and LayerNorm~\cite{ba2016layer}) can be used before the residual connection, and we drop it here for clarity.

\vspace{-0.05in}
\subsection{Decoupled Spatial and Temporal Self-Attention}
\label{sec:DNL}

\begin{figure}[t!]
    \centering
    \includegraphics[width=0.92\linewidth]{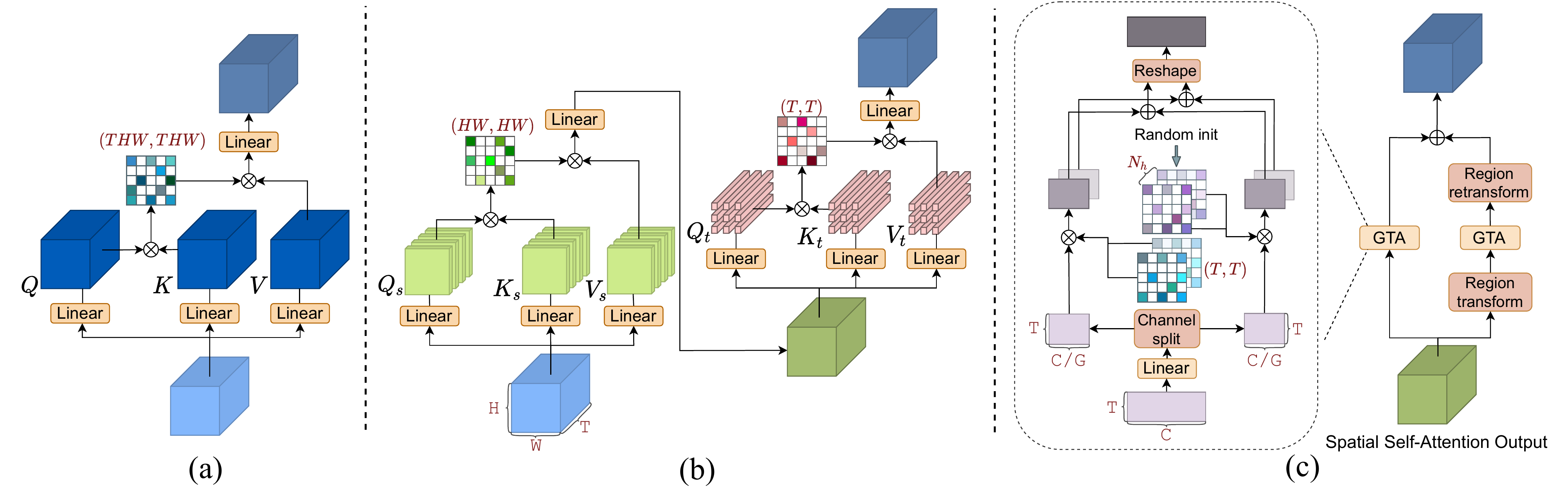}
    \vspace{-2mm}
    \caption{(a) \textbf{Standard self-attention for action recognition}, which computes pairwise similarities between a pixel (query) with other pixels (key) in the spacetime domain. (b) \textbf{Decoupled spatial and temporal self-attention}, which uses separated key/query/value representations for spatial and temporal attention and aggregates spatial and temporal context in a separate manner.
    (c) \textbf{Global temporal attention}, which learns two randomly initialized global attention maps at the pixel-level and the region-level, respectively.
    Regions are derived automatically with a learned transformation matrix.
    Inside the rectangular (dashed line), the spatial dimension of feature maps is omitted. \system is also applied in a cross-channel multi-head fashion, where feature maps are split along the channel dimension into G groups (only 2 groups are shown for simplicity). Residual connections are omitted here for simplicity. See texts for more details.\vspace{-0.05in}}
    \label{fig:model}
\end{figure}

Although self-attention has been widely used in action recognition for capturing spatio-temporal dependencies, we argue in this paper that the coupled modeling of spatial and temporal self-attention prevents the model from learning effective temporal attention. First, when sharing the same transformation matrices for key, query and value, it fails to differentiate between spatial and temporal contexts. This is unsatisfactory for temporal modeling as we need to consider temporal structures of videos instead of simply computing the salient regions by performing self-attention in the spatial domain.
Moreover, when the two attentions are modeled and aggregated jointly, the combined attention tends to be biased towards the appearance similarity as the temporal attention is dominated by the spatial one (see Figure~\ref{fig:teaser}). Based on this observation, we propose the decoupled spatial and temporal self-attention in Figure~\ref{fig:model} (b), which breaks down the standard self-attention block into a spatial self-attention block followed by a temporal self-attention block. We will provide a more in-depth analysis of the decoupled self-attention design via experiments in Section~\ref{sec:results}.

Formally, given the input feature map $X$, we first obtain the three projections through:  $Q_s = XW_Q^s,  K_s = XW_{K}^s, V_s = XW_{V}^s$, where the subscript/superscript $s$ is used to differentiate from the functions in temporal attention.
We then perform the space-only attention on all spatial positions within each frame:
\begin{equation}
    A_s(t) = \text{softmax}\left(\frac{Q_s(t)K_s(t)^T}{\sqrt{C}}\right)V_s(t),
    \quad     Y_s(t) = X(t) + A_s(t)W_{O}^s,
\end{equation}
where $t\in \left\{1, ..., T\right\}$ denotes different time steps.
The output of the spatial attention block is then used as input to the temporal attention block, which performs similar attention operations yet in time only:
\begin{equation}
    A_t(i, j) = \text{softmax}\left(\frac{Q_t(i, j)K_t(i, j)^T}{\sqrt{C}}\right)V_t(i, j), \quad Y_t(i, j) = Y_s(i, j) + A_t(i, j)W_{O}^t,
\label{eq:temporal_att}
\end{equation}
where $i\in \left\{1, ..., H\right\}, j\in \left\{1, ..., W\right\}$ denote different spatial positions, and $Q_t = Y_s W_Q^t,  K_t = Y_s W_{K}^t, V_t = Y_s W_{V}^t$.
Note that although the idea of processing spatial
and temporal information separately has been explored before
for video understanding~\cite{tran2018closer,song2017end}, the effect of disentangling the two dimensions in self-attention is \textit{unknown} in prior work.

\subsection{Global Temporal Attention}
\label{sec:GTA}
We now introduce \system, which is built upon the decoupled self-attention framework and advances the temporal attention to a more effective design. \system aims to learn a global attention map that considers temporal structures and generalizes well for all samples. 

Formally, given the input feature map $X \in \mathbb{R}^{T \times HW \times C}$ generated by the spatial self-attention block, \system models temporal relationships at two different levels of spatial granularity: \textit{pixel-level} and \textit{region-level}. For Pixel \system, all positions in the spatial domain (\ie, $HW$) are treated individually as different samples and temporal modeling is performed along the time axis $T$. As for Region \system, we first project the spatial domain to $K$ semantic regions at each time step $t$. This is achieved by grouping similar pixels with related semantic meanings into the same region~\cite{chen2019graph}: $X_G(t) = G_R(t) X(t)$, where the region transformation matrix $G_R(t) = W_GX(t)^T$ and $W_G \in \mathbb{R}^{K\times C}$ is a learnable weight matrix. Then, temporal modeling is performed across frames on each region individually in the same manner as Pixel GTA, followed by a transposed region transformation matrix $G_R^T$ to reproduce the pixel-level spatial domain.
Similar to Eqn.~\ref{eq:output}, the output of \system can be written as:
\begin{equation}
    Y = X + \underbrace{A_P W_P^O}_{\text{Pixel GTA}} + \underbrace{A_R W_R^O}_{\text{Region GTA}}.
\end{equation}

Unlike conventional self-attention where the attention map is produced by pairwise dot-product interactions (Eqn.~\ref{eq:dotproduct}), we train attention maps that do \textit{not} depend on individual pixel relationships. In particular, we directly learn randomly initialized weight matrices $\hat{M}_P, \hat{M}_R \in \mathbb{R}^{T\times T}$ to modulate the value representation of Pixel and Region \system, respectively:
\begin{align}
    A_P=\hat{M_P}V_P, \quad A_R=G_R^T (\hat{M_R}V_R),
\label{eq:gta_output}
\end{align}
The idea of using a learned global attention matrix rather than pairwise dot product is that pairwise interactions fluctuate across different samples, lacking a global temporal consistency at the dataset level. 
In addition, the standard self-attention fails to consider the ordering of sequences~\cite{DBLP:conf/iclr/CordonnierLJ20}---if we shuffle the pixels used to compute the attention map (\ie Eqn.~\ref{eq:dotproduct}), the attention value between a pair would still be the same in the matrix, thus the output will not change, which is not what we desire.

\vspace{-0.1in}
\paragraph{Cross-channel Multi-head GTA.} The attention matrix $\hat{M}$ in Eqn.~\ref{eq:gta_output} is used to learn a linear combination of $V \in \mathbb{R}^{T \times C}$~\footnote{We omit the subscripts $P$ and $R$ for $A$, $\hat{M}$ and $V$, as the same operations are applied to both Pixel and Region \system. $HW$ are considered as different samples and we omit it for brevity\label{fn:omit}.} across different time steps, without considering feature interactions in the channel dimension. We further improve temporal modeling by incorporating channel interactions. We split $C$ into $G$ groups, and for each group and we apply a multi-head \system. In particular, for the $g$-th group, the outputs of the multi-head attention $\texttt{MH}_g$ is:
\vspace{-0.1in}
\begin{align}
    \texttt{MH}_g = \texttt{Concat}_{k=1}^{N_h}(\hat{M}^k_g V_g) \in \mathbb{R}^{N_h \times T \times \floor*{\frac{C}{G}} }, 
\end{align}

where $\hat{M}^k_g \in \mathbb{R}^{T \times T}$ represents the $k$-th head for the $g$-th group, $V_g \in \mathbb{R}^{T \times \floor*{\frac{C}{G}}}$ denotes the value for the $g$-th group and $N_h$ denotes the number of heads used. Each head focuses on distinct temporal attention patterns. To capture interactions across different groups, we sum the outputs along the channel dimension between different groups to produce $\texttt{MH}_G$ as:
\vspace{-0.1in}
\begin{align}
     \texttt{MH}_G = \sum_{g=1}^{G} \texttt{MH}_g \in \mathbb{R}^{N_h \times T \times  \floor*{\frac{C}{G}}},
\end{align}
which mixes information across channels in different groups. In order for $\texttt{MH}_G$ to have the same size as $X \in \mathbb{R}^{T \times C}$~\footref{fn:omit} for residual addition, one can transform $\texttt{MH}_G$ with an additional layer. Instead, we simply set $N_h$ to be $G$ and reshape $\texttt{MH}_G$ to be the same size of $\mathbb{R}^{T \times C}$.

\begin{figure}[t]
\begin{floatrow}
\capbtabbox{
    \centering
    \scriptsize
    \setlength{\tabcolsep}{4pt}
    \renewcommand{\arraystretch}{1.2}
        \begin{tabular}{@{\extracolsep{\fill}\;}l|cc|cc@{\extracolsep{\fill}\;}}
            \toprule
             \multicolumn{1}{l|@{}}{\bf Model} & {\scriptsize \bf FLOPs} & {\scriptsize \bf \#Params} & {\bf SSv1} & {\bf SSv2}  \\
            \midrule
             R2D-50 & 32.7 G & 23.9 M & 17.0 & 26.8 \\
              \quad + NL & 61.1 G & 31.2 M & 31.2 & 50.7 \\
              \quad + DNL & 49.9 G & 31.2 M & 38.8 & 55.5 \\
              \quad + \system & 50.2 G & 31.2 M & \textbf{50.6} & \textbf{63.5} \\
            \midrule
             SlowFast-R50 & 131.4 G & 34.0 M & 50.9 & 63.4 \\
              \quad + NL & 239.9 G & 41.4 M & 51.7 & 63.9 \\
              \quad + DNL & 169.1 G & 41.4 M & 52.0 & 64.1 \\
              \quad + \system & 169.9 G & 41.4 M & \textbf{53.4} & \textbf{64.9}  \\
            \bottomrule
        \end{tabular}
    \vspace{-2mm}
}{
    \caption{Compare \system with the standard / decoupled non-local block (NL / DNL). \vspace{-6mm}}
    \label{tab:compare_original}
}
\capbtabbox{
    \centering
    \scriptsize
    \setlength{\tabcolsep}{3pt} %
    \renewcommand{\arraystretch}{1.2}
        \begin{tabular}{@{\extracolsep{\fill}\;}lccc@{\extracolsep{\fill}\;}}
            \toprule
            \multicolumn{1}{c@{}}{\textbf{Method}} & \textbf{GFLOPs$\times$views} & \textbf{Top-1} & \textbf{Top-5}\\
            \midrule
            TSM~\cite{lin2019tsm} & 86$\times$30 &  74.7 & 91.4 \\
            bLVNet-TAM~\cite{fan2019more} & 93$\times$9 & 73.5 & 91.2 \\
            MSNet~\cite{kwon2020motionsqueeze} & 87$\times$10 & 76.4 & - \\
            S3D-G~\cite{xie2018rethinking} & 143$\times$N/A &  77.2 & 93.0 \\
            I3D+NL~\cite{Wang2017NonlocalNN} & 359$\times$30 &  77.7 & 93.3 \\
            CorrNet-R101~\cite{wang2020video} & 224$\times$30 & 79.2 & - \\
            \midrule
            R2D-R50 + NL & 77$\times$30 & 74.8 & 91.5 \\
            R2D-R50 + \system & 62$\times$30 & 75.9 & 92.2 \\
            SlowFast-R101 + NL & 137$\times$30 &  78.9 & 93.9 \\
            \textbf{SlowFast-R101 + \system} & 137$\times$30 & \textbf{79.8}& \textbf{94.1}\\
            \bottomrule
        \end{tabular}
    \vspace{-2mm}
}{
    \caption{Comparisons with state-of-the-art methods on Kinetics-400 dataset. }%
        \label{tab:stoa-kinetics}
}
\end{floatrow}
\end{figure}

%% file: experiment.tex
\vspace{-0.1in}
\section{Experiments}
\vspace{-0.05in}

We extensively evaluate our approach on three video action benchmarks, including two temporal-related datasets: Something-Something (v1\&v2)~\cite{goyal2017something}, and a large-scale dataset that is less sensitive to temporal relationships: Kinetics-400 (K400)~\cite{Carreira_2017}. 
As we aim to improve temporal modeling for video action recognition, our experiments focus more on temporal sensitive datasets (SSv1 and SSv2).
\system is flexible and can be easily inserted into existing 2D and 3D backbones.
In our experiments, we adopt the standard R2D-50 network~\cite{he2016deep} and the SlowFast-R50 network~\cite{feichtenhofer2019slowfast} as our 2D/3D backbones.
More dataset-specific training and testing details are available in the supplementary material.

\vspace{-0.1in}
\subsection{Main Results}
\label{sec:results}

\paragraph{Effectiveness of \system in a decoupled framework.} We report the results of \system using both 2D and 3D backbones and compare with the alternative approaches: (1) standard non-local block (\textsc{NL})~\cite{Wang2017NonlocalNN}, which is a variant of self-attention that flattens all pixels in space and time dimension into a huge vector; (2) decoupled non-local block (\textsc{DNL}), which breaks down NL into spatial self-attention followed by temporal self-attention. For both of our approaches and the compared baselines, we apply five blocks (2 to $\text{res}_3$ and 3 to $\text{res}_4$ for every other residual block) in the backbone networks unless specified, following \cite{Wang2017NonlocalNN}.

Table~\ref{tab:compare_original} summarizes the comparison results. We first observe a huge gap between the performance of 2D and 3D backbones, which shows the importance of utilizing temporal information for SSv1\&SSv2 datasets.
Notably, we see that by simply separating temporal self-attention from spatial self-attention, \textsc{DNL} outperforms \textsc{NL} on both backbones, while requiring 20\%-30\% less computation cost. Compared to \textsc{NL}, \textsc{DNL} offers a 7.6\% / 4.8\% gain on SSv1 / SSv2 in the 2D setting.
This suggests that the spatial and temporal self-attentions should be treated \textit{separately} to capture more informative temporal contexts.
Finally, \system produces the best results on the two datasets with both 2D and 3D backbones with reduced FLOPs comparing to NL. For example, on the 2D backbone, \system further outperforms \textsc{DNL} by 11.8\% / 8.0\% on SSv1, SSv2, respectively, confirming the effectiveness of \system for temporal modeling. On a 3D backbone, we observe similar trends with gains. This highlights the compatibility of \system for both 2D and 3D networks.
It is also noteworthy that 2D networks can achieve comparable performance with 3D backbones when equipped with \system.

\begin{table}[t!]
    \caption{Comparisons with state-of-the-art methods on Something-Something v1 \& v2 datasets. Top-1 and Top-5 accuracy on validation set are reported here. \textbf{Bold} and \underline{underline} shows the highest and second highest results.\vspace{-0.15in}}
    \label{tab:stoa-ss}
    \centering
    \renewcommand{\tabcolsep}{4pt}
    \scriptsize
        \begin{tabular}{@{\extracolsep{\fill}\;}llcccccc@{\extracolsep{\fill}\;}}
            \toprule
            \multicolumn{1}{c@{}}{\multirow{2}{1.5cm}{\bf Method}} & \multirow{2}{1.5cm}{\textbf{Backbone}} & \multirow{2}{1cm}{\textbf{Pretrain}} & \multirow{2}{2cm}{\textbf{Frames$\times$Crops \\
            ~~~~~~~$\times$Clips}} & \multicolumn{2}{c}{SSv1} & \multicolumn{2}{c@{}}{SSv2} \\
            \cmidrule{5-6} \cmidrule{7-8} 
            & & & & \textbf{Top-1} & \textbf{Top-5} & \textbf{Top-1} & \textbf{Top-5}\\
            \midrule
            TRN~\cite{zhou2018temporal}& BNInception& ImgNet & 8$\times$1$\times$1 &34.4 &- &48.8 &-\\
            TSM~\cite{lin2019tsm}& 2D R50& ImgNet & 8$\times$1$\times$1 &45.6 &74.2 &58.8 &85.4\\
            TSM~\cite{lin2019tsm}& 2D R50& ImgNet & 16$\times$1$\times$1 &47.3 &77.1 &61.2 &86.9\\
            $\text{TSM}_\text{RGB+Flow}$~\cite{lin2019tsm}& 2D R50& ImgNet & (16+16)$\times$1$\times$1 &52.6 &81.9 &65.0 &89.4\\
            MSNet~\cite{kwon2020motionsqueeze}& 2D R50+TSM& ImgNet & 8$\times$1$\times$1 &50.9 &80.3 &63.0 &88.4\\
            MSNet~\cite{kwon2020motionsqueeze}& 2D R50+TSM& ImgNet & 16$\times$1$\times$1 &52.1 &82.3 &64.7 &89.4\\
            $\text{MSNet}_{En}$~\cite{kwon2020motionsqueeze}& 2D R50+TSM& ImgNet & (16+8)$\times$1$\times$10 &\underline{55.1} &\textbf{84.0} &\underline{67.1} &\underline{91.0}\\
            \midrule
            ECO~\cite{zolfaghari2018eco}& 3D R18+BNInc &K400 & 16$\times$1$\times$1 &41.4 &- &- &-\\
            $\rm ECO_{En}\emph{Lite}$~\cite{zolfaghari2018eco}& BNInc+3D R18 &K400 &92$\times$1$\times$1 &46.4 &- &- &-\\
            I3D+NL~\cite{Wang2017NonlocalNN}& 3D R50& K400& 32$\times$3$\times$2 &44.4 &76.0 &- &-\\
            I3D+NL+GCN~\cite{wang2018videos}& 3D R50& K400& 32$\times$3$\times$2 &46.1 &76.8 &- &-\\
            S3D-G~\cite{xie2018rethinking} & 3D Inception & ImgNet& 64$\times$1$\times$1 & 48.2 &78.7 &- &-\\
            CorrNet~\cite{wang2020video} & 3D CorrNet-50 & - & 32$\times$1$\times$10 & 48.5 &- &- &-\\
            CorrNet~\cite{wang2020video} & 3D CorrNet-101 & - & 32$\times$3$\times$10 & 51.1 &- &- &-\\
            TEA~\cite{li2020tea}& 3D R50& ImgNet & 8$\times$1$\times$1 &48.9 &78.1 &- &-\\
            TEA~\cite{li2020tea}& 3D R50& ImgNet & 16$\times$3$\times$10 &52.3 &81.9 &65.1 &89.9\\
            \midrule
            \midrule
            \system & 2D R50 & ImgNet & 8$\times$1$\times1$ & 50.6& 78.8 &63.5 &88.6\\
            \system & 2D R50 & ImgNet & 16$\times$1$\times1$ & 52.0& 80.5 &64.7 &89.3\\
            \system & 2D R50+TSM & ImgNet & 8$\times$1$\times1$ & 51.6& 79.8 &63.7 &88.9\\
            \system & 2D R50+TSM & ImgNet & 16$\times$1$\times1$ & 53.7& 81.7 &65.3 &89.6\\
            $\textbf{\system}_{En}$ & 2D R50+TSM & ImgNet & (16+8)$\times$3$\times2$ & \textbf{56.5}&  \underline{83.1} & \textbf{68.1} & \textbf{91.1}\\
            \bottomrule
        \end{tabular}
    \vspace{-0.1in}
\end{table}

\vspace{-0.05in}
\subsection{Comparison with State-of-the-art}
\paragraph{Kinetics-400} Table~\ref{tab:stoa-kinetics} presents the comparative results with other state-of-the-art methods on Kinetics-400. The first section of the table shows the methods based on 2D CNN network. The second section contains the models with 3D CNN backbone. The third section illustrates the comparison of our \system and NL added to 2D and 3D CNN backbones. We can see that \system achieves consistent improvement over the NL counterpart on 2D and 3D CNN backbones. And adding \system to SlowFast-R101 can achieve 79.8\% top-1 accuracy on Kinetics-400 dataset, which is the state-of-the-art performance on Kinetics-400.

\vspace{-0.05in}
\paragraph{Something-Something v1\&v2} We also compare our approach with the state-of-the-art methods on SSv1 \& SSv2 datasets.
As shown in Table~\ref{tab:stoa-ss}, given 8 input frames, our approach based on 2D RestNet-50 with TSM backbone achieves 51.6\% and 63.7\% on SSv1 and SSv2 at top-1 accuracy, respectively.
Specifically, with the same number of input frames, our approach outperforms TRN~\cite{zhou2018temporal} which utilizes relation networks, and MSNet~\cite{kwon2020motionsqueeze} which incorporates the motion features. This demonstrates that our proposed \system is more effective in temporal modeling.
Our approach also achieves superior results when compared with the recent work that leverages additional modules to improve 3D CNN backbones, such as the non-local block (I3D+NL~\cite{Wang2017NonlocalNN}), GCN (I3D+NL+GCN~\cite{wang2018videos}), the correlation operation (CorrNet~\cite{wang2020video}), and the multiple temporal aggregation module and the motion excitation module (TEA~\cite{li2020tea}). 
Finally, we evaluate the ensemble model ($\text{\system}_{En}$) by averaging output prediction scores of the 8-frame and 16-frame models, and obtain 56.5\% and 68.1\% at top-1 accuracy on SSv1 and SSv2, respectively, which achieves the state-of-the-art performance.

\thisfloatsetup{floatrowsep=myfill}
\begin{figure}[t]
\begin{floatrow}
\capbtabbox{
    \centering
    \scriptsize
    \setlength{\tabcolsep}{1.7pt}
    \renewcommand{\arraystretch}{1.2}
        \begin{tabular}{@{\extracolsep{\fill}\;}ccc|cc@{}}
            \toprule
             \textbf{Pixel} & \textbf{Region} & \textbf{CCMH} & \textbf{Top-1} & \textbf{$\bigtriangleup$}  \\
             \midrule
            \checkmark & \checkmark & \checkmark & 50.6 & --\\
            \midrule
            \checkmark &  & \checkmark & 49.6 & -1.0 \\
             & \checkmark & \checkmark & 47.9 & -2.7 \\
            \checkmark & \checkmark &  & 49.4 & -1.2 \\
            \checkmark &  &  & 49.1 & -1.5 \\
             & \checkmark &  & 46.3 & -4.3 \\
            \bottomrule
        \end{tabular}
        \vspace{-2mm}
}{
    \caption{Contribution of different components in \system. }
    \label{tab:components}
}
\capbtabbox{
    \centering
    \scriptsize
    \setlength{\tabcolsep}{1.5pt} %
    \renewcommand{\arraystretch}{1.2}
        \begin{tabular}{@{\extracolsep{\fill}\;}l|cc}
            \toprule
             \multicolumn{1}{c@{}|}{\textbf{Model}}& \textbf{Original} & \textbf{Decoupled}\\
            \midrule
            \system  & -- &  \textbf{50.6}\\
            \midrule
            NL~\cite{Wang2017NonlocalNN} & 31.2 & 38.8 \\
            CGNL~\cite{yue2018compact} & 26.7 & 37.4 \\
            GCNet~\cite{cao2019gcnet} & 28.4 & 39.0 \\ 
            GloRe~\cite{chen2019graph} & 33.2 & 38.6 \\
            \bottomrule
        \end{tabular}
        \vspace{-2mm}
}{
    \caption{Comparisons with recent NL variants.}
    \label{tab:compare_variants}
}
\capbtabbox{
    \centering
    \scriptsize
    \setlength{\tabcolsep}{1.3pt} %
    \renewcommand{\arraystretch}{1.2}
        \begin{tabular}{@{\extracolsep{\fill}\;}l|cc|cc@{}}
            \toprule
            \multicolumn{1}{c@{}|}{\textbf{Model}} & {\bf FLOPs} & {\bf \#Params}  &\textbf{Top-1} & \textbf{$\bigtriangleup$}\\
            \midrule
            R2D-50 & 32.7 G & 23.9 M& 17.0 & - \\
            \midrule
            +SA & 41.7 G & 27.5 M& 17.9 & +0.9 \\
            +TA & 41.0 G& 27.5 M& 37.6 & +20.6 \\
            +SA+TA & 49.9 G& 31.2 M& 38.8 & +21.8 \\
            +SA+TAPE & 49.9 G& 31.2 M& 48.4 & +31.4 \\
            +SA+\system & 50.2 G& 31.2 M& 50.6 & +33.6\\
            \bottomrule
        \end{tabular}
        \vspace{-2mm}
}{
    \caption{Impact of SA, TA and temporal order. }
    \label{tab:blocks}
}
\end{floatrow}
\end{figure}

\thisfloatsetup{floatrowsep=myfill}
\begin{figure}[t!]
\begin{floatrow}
\ffigbox{
    \centering
    \adjincludegraphics[width=\linewidth, trim={0 {0.42\height} {0.22\width} 0},clip]{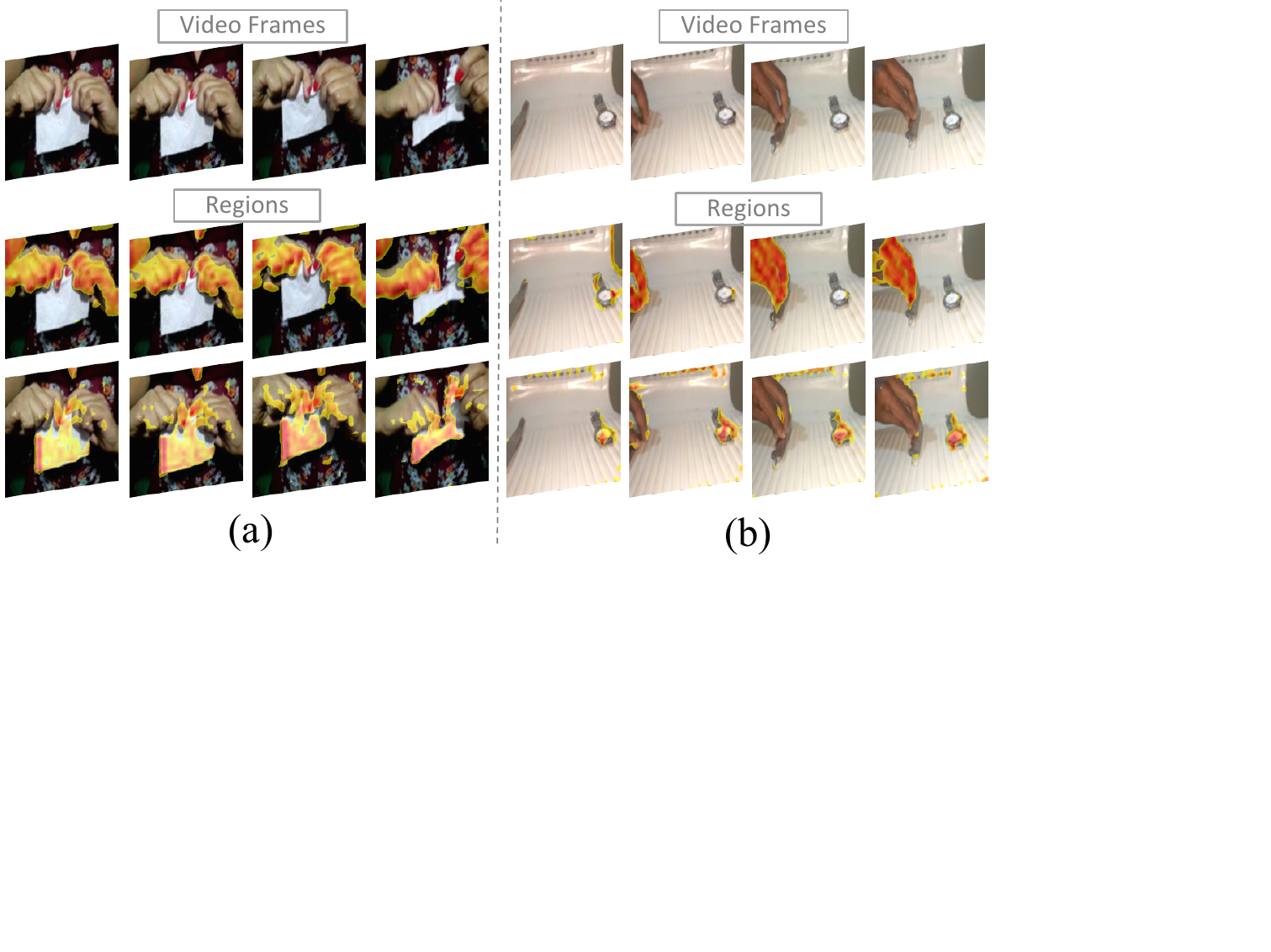}
    \vspace{-0.1in}
}
{
    \caption{Regions visualization: (a)``Tearing smth. into two pieces"; (b)``Moving smth. closer to smth.". The second and third rows are regions obtained by Region \system.\vspace{-0.1in}}
    \label{fig:proj_weight}
}
\ffigbox{
    \centering
    \includegraphics[width=0.9\linewidth]{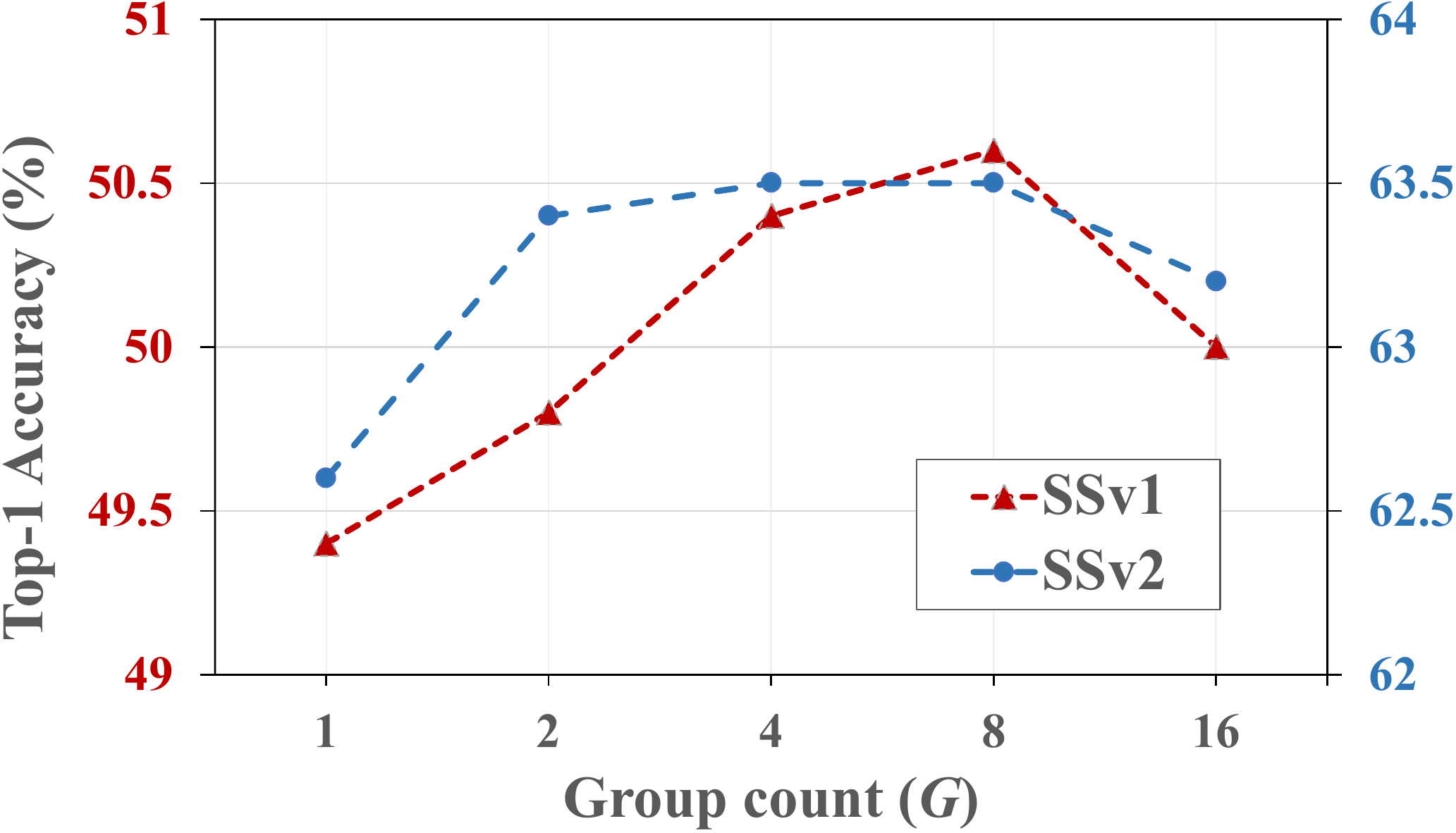}
    \vspace{-0.1in}
}
{
    \caption{Impact of group count. Top-1 accuracy on Something-Something v1\&v2 datasets are reported here.\vspace{-0.1in}}
    \label{fig:group_count}
}
\end{floatrow}
\end{figure}

\subsection{Ablative Studies}
\label{sec:ablation}
We conduct extensive ablation studies on SSv1 using R2D backbone. More analysis are available in the supplementary material.

\vspace{-0.1in}
\paragraph{Contribution of Different Components.} 
We first validate the contribution of each component in \system by removing them from the full model.
As shown in Table~\ref{tab:components}, while Pixel \system plays a more important role than Region \system, the combination of these two modules yields the best result, achieving more than 1\% improvement compared to using each of them alone. It indicates that Pixel \system and Region \system are \textit{complementary} to each other, focusing on learning temporal relationships at different levels of spatial granularity. 
We further visualize regions that are automatically discovered by Region \system in Figure~\ref{fig:proj_weight}. We can see that Region \system is capable of discovering regions that share similar semantic meanings. For example, in the first video, the ``hand'' and the ``paper'' are automatically identified as different regions, while the ``hand" and the ``watch" are detected in the second video.
Table~\ref{tab:components} also shows the contribution of the cross-channel multi-head (CCMH) design when the group size is set to 8. Specifically, CCMH has a larger impact on Region \system than Pixel \system (1\% gain v.s. 0.5\% gain) and we hypothesize that modeling temporal relationships at the region level is more challenging and requires channel interactions. With the improved performance of Region \system, the fusion of pixel-level and region-level information becomes more beneficial when CCMH is applied (1.0\% gain v.s. 0.3\% gain w/o CCMH).

\vspace{-0.1in}
\paragraph{Temporal modeling in NL variants.} 
Recent work has focused on improving the vanilla non-local block by introducing channel-wise attention~\cite{cao2019gcnet,yue2018compact} or graph-based reasoning~\cite{chen2019graph}.
Although these variants have been applied to video action recognition, their capacity to model temporal relations is relatively underexplored.
In Table~\ref{tab:compare_variants}, we provide a side-by-side comparison with these NL variants and their decoupled version on SSv1.
We first observe that all three variants fail to achieve satisfying improvements over the vanilla NL (31.2\%). In particular, the use of extra channel-wise attention ( CGNL~\cite{yue2018compact}, GCNet~\cite{cao2019gcnet}) leads to even worse results, indicating that the entangled modeling of spatial, temporal and channel interactions in fact hinders the learning of temporal relationships. Interestingly, by simply decoupling the spatial and temporal operations, substantial improvements can be achieved for all three variants and the results are comparable with DNL (38.8\%). 
Nevertheless, our \system outperforms these NL variants by clear margins, which demonstrates its superior capacity to model temporal information.

\vspace{-0.1in}
\paragraph{Miscellaneous.}
In Table~\ref{tab:blocks}, we compare the contribution of spatial and temporal self-attention modules, as well as the impact of modeling temporal order in temporal self-attention. As the SSv1 dataset relies highly on temporal relationships, applying spatial self-attention (\textsc{SA}) alone in the spatial domain slightly improves the backbone network (0.9\% gain). In contrast, using the temporal self-attention (\textsc{TA}) provides much more significant improvements (20.6\% gain). Adding positional encoding to the temporal self-attention module (\textsc{TAPE}) further improves the performance by 9.6\%, which proves the importance of modeling temporal order information.
Finally, our \system achieves the best result with a negligible increase in computation cost. It is worth noting that our Pixel \system (without applying \system to regions) already outperforms TAPE no matter whether CCMH is used or not (49.1\% / 49.6\% in Table~\ref{tab:components}). This verifies that our \system design is more effective in temporal modeling than temporal self-attention and positional encoding.

We also evaluate different values of group count used in \system in Figure~\ref{fig:group_count}. We can see that using a group count larger than 1 can largely improve the performance, which demonstrates the importance of channel interactions in \system. And a group count of 8 offers the best performance on SSv1 and SSv2. When the group count becomes larger than 8, the performance drops because the number of channels in each group becomes too small.

%% file: appendix.tex
\appendix
\section*{Appendix}

Section~\ref{sec:testing_results} reports additional results on the test set of Something-Something v1\&v2.
Section~\ref{sec:more_ablative} presents more ablative study results of \system. 
Section~\ref{sec:relation} elaborates on \system that is designed for better temporal modeling.
Section~\ref{sec:more_visualization} shows more visualization results of global temporal attention weights, transformed regions and swapped attention.
Finally, Section~\ref{sec:more_setups} provides dataset-specific implementation details on training and testing.

\section{Testing Results on Something v1\&v2}
We compare the performance of our approach on the test set with the state-of-the-art methods on Something-Something v1 \& v2 datasets.
As is shown in Table~\ref{tab:stoa-ss}, our approach based on 2D RestNet-50 with TSM backbone achieves 49.8\% and 66.9\% on SSv1 and SSv2 at top-1 accuracy, respectively. Although on SSv1 dataset, it is still below the $\text{TSM}_\text{RGB+Flow}$, $\text{TSM}_\text{RGB+Flow}$ is based on the two-stream network and utilizes additional optical flow information. With only RGB input, our \system achieves the best performance among the recently proposed STM~\cite{jiang2019stm} and bLVNet-TAM~\cite{fan2019more} on 2D CNN backbone; I3D+NL+GCN~\cite{wang2018videos} and TEA~\cite{li2020tea} on 3D CNN backbone.
\label{sec:testing_results}
\begin{figure}[h]
\begin{floatrow}
\capbtabbox{
    \centering
    \scriptsize
    \setlength{\tabcolsep}{1.2pt}
    \renewcommand{\arraystretch}{1.2}
    \begin{tabular}{@{\extracolsep{\fill}\quad}llccc}
        \toprule
        \textbf{Method} & \textbf{Backbone} & \textbf{Frames} & \textbf{SSv1} & \textbf{SSv2} \\
        \midrule
        $\text{TRN}_\text{RGB+Flow}$~\cite{zhou2018temporal}& BNInc & 8+8  &40.7 &56.2 \\
        TSM~\cite{lin2019tsm}& 2D R50 & 16  &46.0 &64.3\\
        $\text{TSM}_\text{RGB+Flow}$~\cite{lin2019tsm}& 2D R50 & 16+16 &\textbf{50.7} &\underline{66.6}\\
        STM~\cite{jiang2019stm}& 2D R50 & 16 &43.1 &63.5\\
        bLVNet-TAM~\cite{fan2019more}& 2D R101 & 64 &48.9 &-\\
        \midrule
        $\rm ECO_{En}\emph{Lite}$~\cite{zolfaghari2018eco}& BNInc+3D R18 &92 &42.3 &- \\
        I3D+NL+GCN~\cite{wang2018videos}& 3D R50& 32 &45.0 &- \\
        TEA~\cite{li2020tea}& 3D R50 & 16 &46.6 &63.2\\
        \midrule
        \midrule
        $\textbf{\system}_{En}$ & 2D R50+TSM & 16+8 & \underline{49.8}&  \textbf{66.9}\\
        \bottomrule
    \end{tabular}
}{
    \caption{Results on the test set of Something-Something v1 \& v2 datasets.} %
    \label{tab:stoa-ss}
}
\end{floatrow}
\end{figure}

\section{More Ablative Studies}
\label{sec:more_ablative}

\paragraph{Impact of inserting positions and number of blocks}
Table~\ref{tab:stage} explores the performance of different inserting positions and the number of blocks inserted. We see that even a single \system block inserted at $\text{res}_3$ or $\text{res}_4$ can bring significant improvement over the baseline. However, the enhancement on $\text{res}_5$ is relatively minor. We hypothesize that the final residual stage loses too much fine-grained spatial information, which may hinder the learning of temporal attention at the pixel-level and the region-level. Following the common practice~\cite{Wang2017NonlocalNN}, our full model inserts five \system blocks to leverage the complementary information provided by different residual stages and achieves the best result.

\paragraph{Comparison with Temporal Attention with Positional Embedding (TAPE)}
\label{sec:PE}
Our GTA module is more effective in temporal modeling than TAPE because it not only considers the chronological order of video frames but also models the temporal relationships among them.
Results in Table 5 of the main paper show that GTA outperforms TAPE by \textbf{2.2\%} on SSv1.
Here, we provide a side-by-side comparison between TAPE and our Pixel GTA (without applying GTA to regions) in Table~\ref{tab:pe}. Our Pixel GTA consistently outperforms TAPE under different settings. Furthermore, TAPE can also benefit from our cross-channel multi-head (CCMH) design, but Pixel GTA still achieves the best performance.

\paragraph{Impact of number of regions.}
\label{sec:region}
We conduct experiments on the impact of the number of regions used in Region\system in Table~\ref{tab:region}. We can see that when increasing the number of regions from $C/32$ to $C$ ($C$ is the channel dimension of the feature map), the accuracy increase first and reach the peak when $K=C/8$. More importantly, our Region\system consistently outperforms the model without Region\system under different values of $K$, which proves the effectiveness of our Region\system design.

\thisfloatsetup{floatrowsep=myfill}
\begin{figure}[t!]
\begin{floatrow}
\capbtabbox{
    \centering
    \scriptsize
    \setlength{\tabcolsep}{8pt} %
    \renewcommand{\arraystretch}{1.1}
        \begin{tabular}{@{\extracolsep{\fill}\;}ccc|c@{\extracolsep{\fill}\;}}
            \toprule
            \textbf{$\text{res}_3$} & \textbf{$\text{res}_4$} & \textbf{$\text{res}_5$} & \ \textbf{Top-1}\ \\
            \midrule
             & & & 17.0 \\
             \midrule
            +1 & & & 46.2  \\
            & +1 &  &  46.4  \\
            & & +1  &  37.4 \\
            +1 & +1 & &  49.5  \\
            +2 & +3 & &  50.6 \\
            \bottomrule
        \end{tabular}
}{
    \caption{Impact of inserting positions and number of blocks.}%
    \label{tab:stage}
}
\hspace{0.4in}
\capbtabbox{
    \centering
    \scriptsize
    \setlength{\tabcolsep}{5pt} %
    \renewcommand{\arraystretch}{1.3}
        \begin{tabular}{@{\extracolsep{\fill}\;}lccc}
            \toprule
            \multicolumn{1}{c}{\textbf{Model}} & \textbf{w/o CCMH} & \textbf{w/ CCMH} \\
            \midrule
            + TAPE & 46.5 & 47.2 \\
            + Pixel GTA & \textbf{48.0} & \textbf{48.5}\\ [0.2em]
            \hline \\ [-1.3em]
            + SA + TAPE & 48.4 & 48.8 \\
            + SA + Pixel GTA & \textbf{49.1} & \textbf{49.6} \\
            \bottomrule
        \end{tabular}
}{
    \caption{Ablation on positional embedding (TAPE) and cross-channel multi-head (CCMH) design.}
    \label{tab:pe}
}
\end{floatrow}
\end{figure}

\thisfloatsetup{floatrowsep=myfill}
\begin{figure}[t!]
\begin{floatrow}
\capbtabbox{
    \centering
    \scriptsize
    \setlength{\tabcolsep}{2pt} %
    \renewcommand{\arraystretch}{1.3}
        \begin{tabular}{@{\extracolsep{\fill}\;}c|ccccccc@{\extracolsep{\fill}\;}}
            \toprule
            \textbf{Number of Regions} & w/o RegionGTA & C & C/2 & C/4 & C/8 & C/16 & C/32\\
             \midrule
             \textbf{Top-1} & 49.6 & 49.7 & 50 & 50.3 & \textbf{50.6} & 50.3 & 50.1 \\
            \bottomrule
        \end{tabular}
}{
    \caption{Impact of number of regions. $C$ denotes the channel dimension of the feature map. Top-1 accuracy on SSv1 validation dataset are reported here.}
    \label{tab:region}
}
\hspace{0.1in}
\capbtabbox{
    \centering
    \scriptsize
    \setlength{\tabcolsep}{8pt} %
    \renewcommand{\arraystretch}{1.3}
        \begin{tabular}{@{\extracolsep{\fill}\;}c|c@{\extracolsep{\fill}\;}}
            \toprule
            \textbf{Model} & \textbf{Top-1} \\
            \midrule
            Cross-channel Multi-head & \textbf{50.6} \\
            Multi-head & 50.1 \\
            \bottomrule
        \end{tabular}
}{
    \caption{Comparison on cross-channel multi-head and multi-head.}
    \label{tab:CCMH}
}
\end{floatrow}
\end{figure}

\paragraph{Comparison on cross-channel multi-head (CCMH) and multi-head.}
\label{sec:CCMH}
In Table~\ref{tab:CCMH}, we compare the performance of cross-channel multi-head and multi-head. We can see that the accuracy drops by 0.5\% when the cross-channel design is removed from CCHM. It demonstrates that the channel interaction is also critical to help improve the accuracy of the action recognition task.

\section{Relations to Prior Work}
\label{sec:relation}
Our proposed decoupled framework and the cross-channel multi-head (CCMH) design are the two key differences between \system and the prior work (GloRe~\cite{chen2019graph}). Specifically, our Region GTA generates semantic regions within each frame and performs temporal modeling on each region \textit{individually} along the time axis. In contrast, when applied to spatio-temporal data, GloRe projects the whole 3D feature maps into semantic groups and models the interactions among them. We argue that this kind of grouping and modeling is not capable of capturing effective temporal relationships across different time steps. 
Moreover, GloRe leverages graph convolution to model node-wise interactions, which only considers information diffusion on each channel. Our \system incorporates channel interactions to further improve temporal modeling, and we show its effectiveness in the experiments.

\section{More Visualizations}
\label{sec:more_visualization}

\begin{figure}[t!]
    \centering
    \vspace{-0.5in}
    \includegraphics[width=0.7\linewidth]{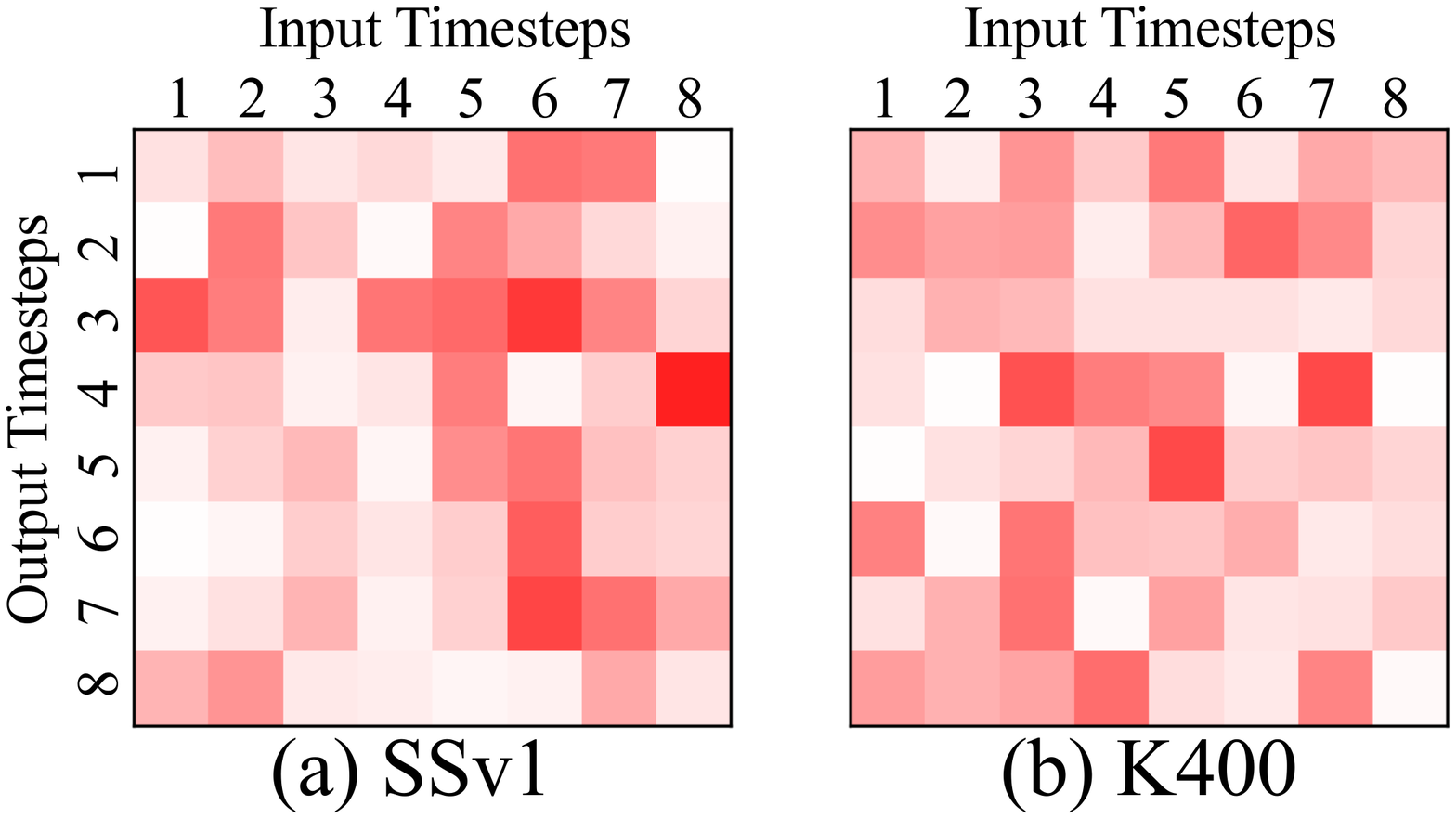}
    \vspace{-0.6in}
    \caption{Visualization of global temporal attention weights on Something-Something v1 and Kinetics-400 datasets. Different columns represent timestamps of input from 1 to 8 and different rows represent timestamps of output from 1 to 8. Darker colors represent larger values of weights.}
    \label{fig:fc_weight}
\end{figure}

\paragraph{Visualization of Global Temporal Attention Weights}
We provide visualization of the global temporal attention weights on two different datasets, Something-Something v1 and Kinetics-400 in Figure~\ref{fig:fc_weight}. 
Specifically, we average the learned global temporal attention weights across different groups and heads, and visualize the absolute value of attention weights. The darker colors represent larger values of weights.
We can see that global attention weights of K400 and SSv1 are visually different. For SSv1, it tends to focus more on the latter part of the frames, while for Kinetics-400, the global temporal attention weights tend to focus more on the middle part of the frames. Our hypothesis is that because there are many action classes "pretending to do something", thus the latter part of the action are of vital importance to distinguish from "pretending to do something" vs "doing something". For example, for "pretending to pick something up" and "picking something up" actions, whether the object has been picked up eventually decides the action type. 
In addition, the global temporal attention weights are not flat across different timestamps, which verifies the effectiveness of our proposed GTA architecture.

\begin{figure}[t!]
    \centering
    \includegraphics[width=0.99\linewidth]{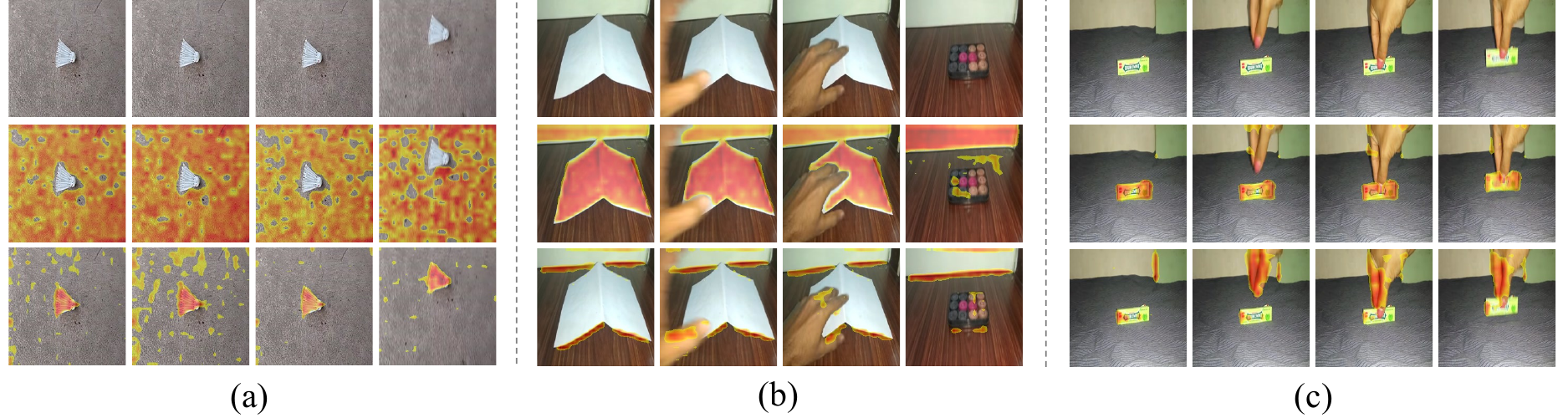}
    \caption{Visualization of the transformed regions of two examples: (a)``Turning the camera downwards while filming something"; (b)``Uncovering something"; (c) ``Picking something up". The first row is the frame sequences. The second and third rows are regions obtained by Region \system.}
    \label{fig:proj_weight_supp}
\end{figure}

\paragraph{Visualization of Transformed Regions}
We present visualization of the transformed regions in Figure~\ref{fig:proj_weight_supp}.
We can see that Region GTA can discover regions that share similar semantic meanings.
For example, in the first video, the ``ground'' region and the ``badminton'' region are automatically identified, the ``paper" and the ``edge" are detected in the second video, and the ``green gum" and the ``hand" are obtained in the third video.

\begin{figure}[t!]
    \centering
    \includegraphics[width=0.99\linewidth]{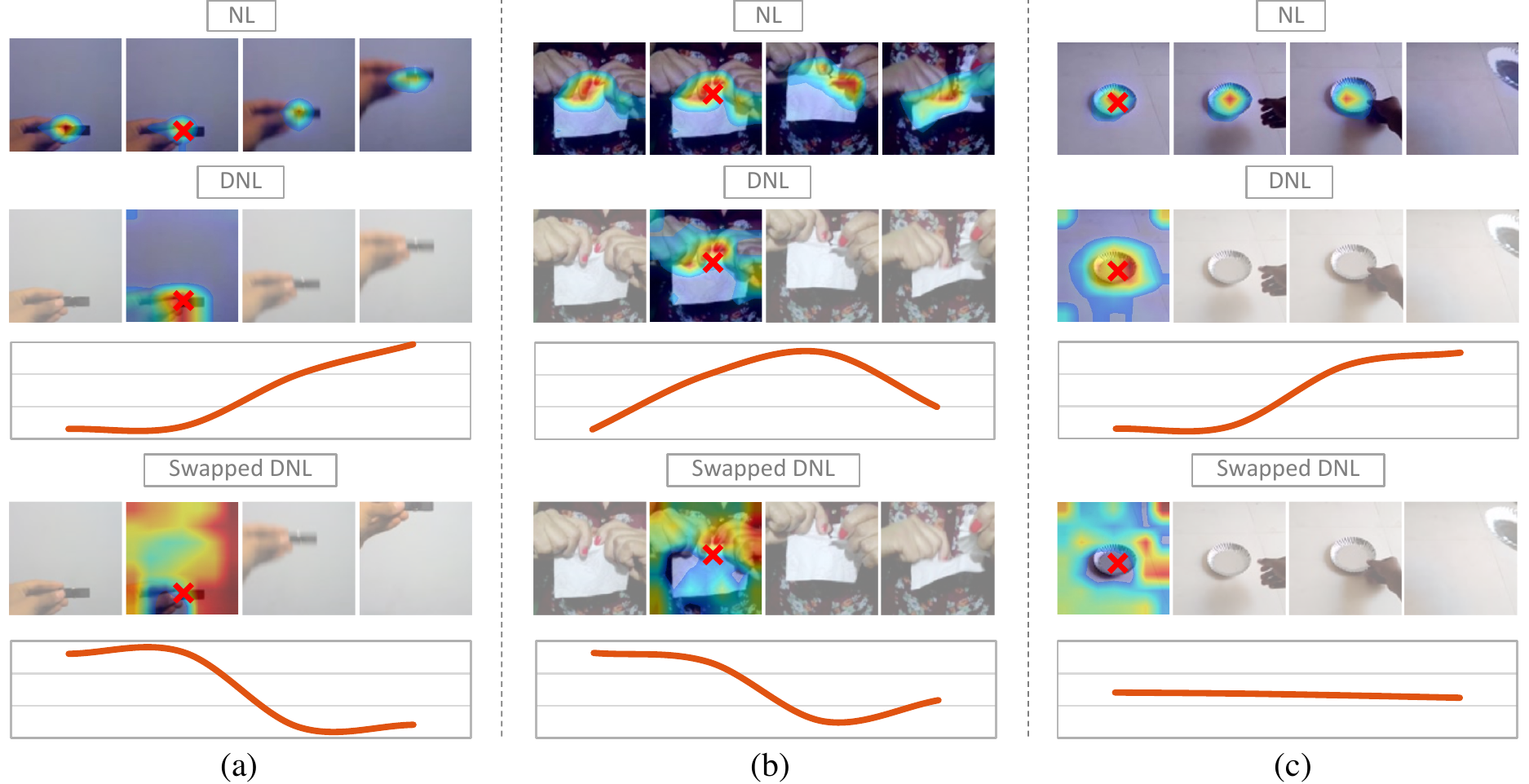}
    \caption{Visualization of the attention maps of three examples: (a)``Moving something up"; (b)``Tearing something into two pieces"; (c)``Picking something up". The first row is the spatio-temporal attention map generated by the non-local module. The second and third row is the spatial and temporal attention map obtained by our decoupled non-local module. The fourth and fifth row is the spatial and temporal attention map generated by swapping the attention functions of the spatial and temporal attention block. The red cross mark denotes the query position.}
    \label{fig:swapped_attention_supp}
\end{figure}

\paragraph{Visualization of Swapped Attention}
To further verify that different context information is needed for spatial and temporal attention, we present the visualization of the swapped attention maps in Figure~\ref{fig:swapped_attention_supp}. Specifically, we swap the attention functions (i.e., query/key/value projections) of the spatial and temporal attention blocks and visualize the attention maps.
We can see that after swapping the spatial and temporal attention functions, the generated temporal attention maps focus more on the frames with similar objects instead of the frames that are useful for recognizing the action.

For example, in Figure~\ref{fig:swapped_attention_supp}(a), the temporal attention weights are larger in the first two frames which share a similar appearance with the same query position (i.e., the pen).
Moreover, the spatial attention maps generated by the temporal attention functions also show substantially different patterns than the original ones. 
The visualization results further verify that different types of context information needed in spatial and temporal attention are captured in the decoupled non-local module.

\section{Experiment Details}
\label{sec:more_setups}

\paragraph{Something-Something v1\&v2~\cite{goyal2017something}}
For the experiments based on the 2D CNN backbone, we follow the same sampling strategy as TSN~\cite{wang2016temporal} to sample 8 frames from the whole video. The same data augmentation is applied as TSN, which first resizes the input frames to 240$\times$320, followed by the multi-scale cropping and random horizontal flipping. Note that we do not flip the clips which include the words ``left" or ``right" in their class labels (e.g., ``pushing something from right to left").
We train the model for 50 epochs and start with a base learning rate of 0.01 with a batch size of 32. The first 2 epochs are used for linear warm-up~\cite{goyal2017accurate} and the learning rate is reduced by a factor of 10 at 30, 40, 45 epochs. The backbone network is initialized with ImageNet pre-trained weights.
For testing, we resize the input images to 240$\times$320 pixels and center crop 224$\times$224 pixels region. We sample 1 clip from each video for the experiments using 2D backbones.

For the experiments based on the 3D CNN backbone, we employ the same training and testing strategy as SlowFast-16$\times$8-R50~\cite{feichtenhofer2019slowfast}. 
We sample 16 and 64 frames for the slow and fast pathways, respectively.

\paragraph{Kinetics-400~\cite{Carreira_2017}}
For the experiments using 2D CNN backbones, we adopt R2D-50 as the backbone and use 8 frames as input. The model is initialized with ImageNet pre-trained weights and trained with step-wise learning schedule following the PySLowFast codebase~\cite{feichtenhofer2019slowfast}. For the experiments using 3D CNN backbones, we use SlowFast-8$\times$8-R101 that samples 8 and 32 frames for the slow and fast pathway, respectively. 
We first train the backbone model on Kinetics-400 and then fine-tune it with GTA, following the same practice for training the non-local blocks~\cite{feichtenhofer2019slowfast}.
We sample 10 clips temporally and 3 crops spatially from each video for testing.